\documentclass{IEEEtran}

\usepackage{amsthm,amsmath}
\usepackage{rotating}
\usepackage[utf8]{inputenc} 
\usepackage{subfigure}

\def\f{\frac}

\def\u{\ifmmode \mathbf{u} \fi}

\def\bi{\begin{itemize}} \def\ei{\end{itemize}}
\def\be{\begin{eqnarray*}}
\def\ee{\end{eqnarray*}}

\def\0{{\mathbf 0}}
\newcommand{\beq}{\begin{equation}}
\newcommand{\eeq}{\end{equation}}

\usepackage{ifthen}
\usepackage{tikz}
\usepackage{multirow}

\usetikzlibrary{calc}

\definecolor{thickLayerColor}{rgb}{.5,.5,.5}
\definecolor{thinLayerColor}{rgb}{.9,.9,.9}

\def\smallImageXPosition{-14em}
\def\largeImageXPosition{0em}
\def\propagationXPosition{-\smallImageXPosition}
\def\layerDist{-1.7}

\definecolor{inColor}{RGB}{150,150,150}
\definecolor{CColor}{RGB}{50,120,200}
\definecolor{PColor}{RGB}{50,200,120}
\definecolor{FColor}{RGB}{50,120,200}
\definecolor{OColor}{RGB}{150,150,150}
\definecolor{concColor}{RGB}{50,200,120}

\tikzstyle{input} = [rectangle, fill=inColor, text width=6em, text centered, minimum height=3em, thick, text = white]
\tikzstyle{C} = [rectangle, fill=CColor, text width=12em, text centered, minimum height=3em, thick, text = white]
\tikzstyle{P} = [rectangle, fill=PColor, text width=12em, text centered, minimum height=3em, thick, text = white]
\tikzstyle{F} = [rectangle, fill=FColor, text width=12em, text centered, minimum height=3em, thick, text = white]
\tikzstyle{O} = [rectangle, fill=OColor, text width=6em, text centered, minimum height=3em, thick, text = white]
\tikzstyle{conc} = [rectangle, fill=concColor, text width=12em, text centered, minimum height=3em, thick, text = white]
\usepackage{multirow}

\begin{document}

\title{Automatic Estimation of Fetal Abdominal Circumference from Ultrasound Images}

\author{Jaeseong~Jang,
        Yejin~Park,
        Bukweon~Kim,
        Sung~Min~Lee,
        Ja-Young~Kwon$^*$,
        and~Jin~Keun~Seo
\thanks{J. Jang is with the Division of Strategic Research,
 National Institute for Mathematical Sciences, Daejeon 34047, South Korea.}
\thanks{B. Kim, S. M. LEE, and J. K. Seo are with the Department of Computational Science and Engineering,
Yonsei University, Seoul 03722, South Korea.}
\thanks{Y. Park and J.-Y. Kwon are with the Department of Obstetrics and Gynecology, Institute of Women’s Life Medical Science, Yonsei University College of Medicine, Seoul 03722, South Korea. Asterisk indicates the corresponding author (e-mail:jaykwon@yuhs.ac.kr).}
}

\maketitle
\begin{abstract}
Ultrasound diagnosis is routinely used in obstetrics and gynecology for fetal biometry, and owing to its time-consuming process, there has been a great demand for automatic estimation. However, the automated analysis of ultrasound images is complicated because they are patient-specific, operator-dependent, and machine-specific. Among various types of fetal biometry, the accurate estimation of abdominal circumference (AC) is especially difficult to perform automatically because the abdomen has low contrast against surroundings, non-uniform contrast, and irregular shape compared to other parameters.We propose a method for the automatic estimation of the fetal AC from 2D ultrasound data through a specially designed convolutional neural network (CNN), which takes account of doctors' decision process, anatomical structure, and the characteristics of the ultrasound image. The proposed method uses CNN to classify ultrasound images (stomach bubble, amniotic fluid, and umbilical vein) and Hough transformation for measuring AC. We test the proposed method using clinical ultrasound data acquired from 56 pregnant women. Experimental results show that, with relatively small training samples, the proposed CNN provides sufficient classification results for AC estimation through the Hough transformation. The proposed method automatically estimates AC from ultrasound images. {  The method is quantitatively evaluated, and shows stable performance in most cases and even for ultrasound images deteriorated by shadowing artifacts. As a result of experiments for our acceptance check, the accuracies are 0.809 and 0.771 with the expert 1 and expert 2, respectively, while the accuracy between the two experts is 0.905.} However, for cases of oversized fetus, when the amniotic fluid is not observed or the abdominal area is distorted, it could not correctly estimate AC.

\end{abstract}

\begin{IEEEkeywords}
fetal ultrasound, fetal biometry, convolutional neural network.
\end{IEEEkeywords}

\section{Introduction}

Ultrasound is the most commonly used tool in the field of obstetrics for the anatomical and functional surveillance of fetuses. Fetal biometry (estimation of the fetal biparietal diameter (BPD), head circumference (HC), and abdominal circumference (AC)) has been known to be useful for predicting intrauterine growth restriction and fetal maturity, and for estimating gestational age \cite{Chalana1996}.
{ 
Acquisition of the standard plane which includes specific anatomical structures as landmarks is prerequisite for the subsequent biometric measurements including BPD, HC, AC, and femur length (FL) \cite{Hadlock1985}. In clinical practice, clinicians manually obtain the standard planes and this process requires knowledge of anatomy and spatial perception thus accuracy is dependent on operator’s experiences \cite{Hao2015, Ni2014}. The accuracy of estimated fetal weight using ultrasound holds intra- and inter-observer variability as the fetal weight is extrapolated from a formulation of fetal biometric measurements \cite{Hadlock1985}. And among biometric measurements, AC is most predictive of fetal weight thus, a variation in AC measurement leads to inaccurate fetal weight estimation \cite{Campbell1975}. To ensure a precise AC plane that is perpendicular to the true fetal longitudinal axis, clinician has to continuously move the transducer to find a plane consisting accurate landmarks. This process is firstly, cumbersome as fetal movement, breathing movement, and fetal position hinder prompt acquisition of the plane; and secondly, may lead to inaccurate measurement as inexperienced operators often fail to adhere to multiple landmarks of correct AC plane \cite{Espinoza2013}.  Therefore, development and implementation of automated fetal biometric measurements has recently gained spotlight in hope to improve clinicians’ workflow and to overcome operator-dependency \cite{Chalana1996, Espinoza2013}.}

{ 
For stable morphologcal information extraction from ultrasound images, numerous methods have been suggested to handle noisy ultrasound images, which are affected by signal dropouts, artifacts, missing boundaries, attenuation, shadows, and speckle \cite{Rueda2014}.
In most methods, in order to deal with such inherent difficulties, image intensity-based or gradient-based methods have been preferred to extract the boundaries of target anatomies \cite{Chalana1996,Pathak2000,YuY2002,Jardim2005,Yu2008}, and abdominal circumference \cite{Wang2014,Nithya2009}.
While the image gradient-based methods shows a stable performance and progresses for HC and FL which have high contrast against surroundings, automatic measurement of AC is considered as more challenging issue because fetal abdomen has low contrast against surroundings, non-uniform contrast, and irregular shape in ultrasound images.

In addition to the boundary extraction methods, it is important to evaluate how a given ultrasound image is proper for AC measurement \cite{Kumar2015}.
Although the evaluation is an essential to automate entire diagnostic process for AC measurement, the model is limited to find a proper plane but not to extract AC.

Instead of such image-based approaches, machine learning methods, such as the probabilistic boosting tree (PBT) \cite{Gustavo2008}, have been used for fetal biometry including AC.
The PBT method is a multi-class discriminative model constructing a tree with its nodes as distinct strong classifiers made by several weak classifiers. By classifying segment structures in ultrasound images, this method estimates fetal biometry parameters \cite{Gustavo2008}. Although this approach showed some notable results, it requires complex, well-annotated data to train the tree.

Recently, with great successes in object recognition, convolutional neural network (CNN) has attracted much attention  and was also applied in fetal biometry to analyze high-level features from ultrasound image data.
Each method aims to find a standard abdominal plane \cite{Hao2015} from succesive ultrasound frames or localize a fetal abdomen \cite{Ravishankar2016} from a ultrasound image.
Although the approaches their function. These approaches, however, implement only a part of an entire measurement process and need to be integrated for full automation of the measurement process.
Additionally, this method has faced obstacles in the clinical environment: (i) it is difficult to collect sufficient data for training, and (ii) it is difficult to cope with serious artifacts including shadowing artifacts \cite{Hao2015}.
}

In this paper, we propose a method that increases classification performance with relatively small number of data and also deals with artifacts by including ultrasound propagation direction as well as multiple scale patches as inputs. The
proposed method classifies images patches from an ultrasound image into anatomical structures so that the classification allows the verification of acceptability of a given abdominal plane.
By detecting anatomical structures in a fetal abdomen, we estimate the AC of the accepted plane using the ellipse detection method based on Hough transform.
We validated our method using ultrasound data of the AC measurement from fetuses at 20-34 weeks of gestation.
Three trained clinicians evaluated the accepted abdominal planes and AC estimated by the method.

The major contribution of our work is as follows:
\begin{itemize}
  \item  We develop a specialized CNN structure that takes account of sonographers' decision process by considering the characteristics of ultrasound imaging.
  The proposed CNN structure shows high training performance in spite of a relatively low number of training samples.
  \item We develop a framework that combines the CNN and Hough transform to complement each other.
  {The CNN simultaneously provides evidence for AC plane evaluation and pre-processing of an ultrasound image for AC estimation.}
  With the combination, we can achieve a more stable AC estimation compared to the case of using a mathematical model alone.
\end{itemize}

\section{Methods}

\begin{figure*}[h!]
\centering
\includegraphics[width=1\textwidth]{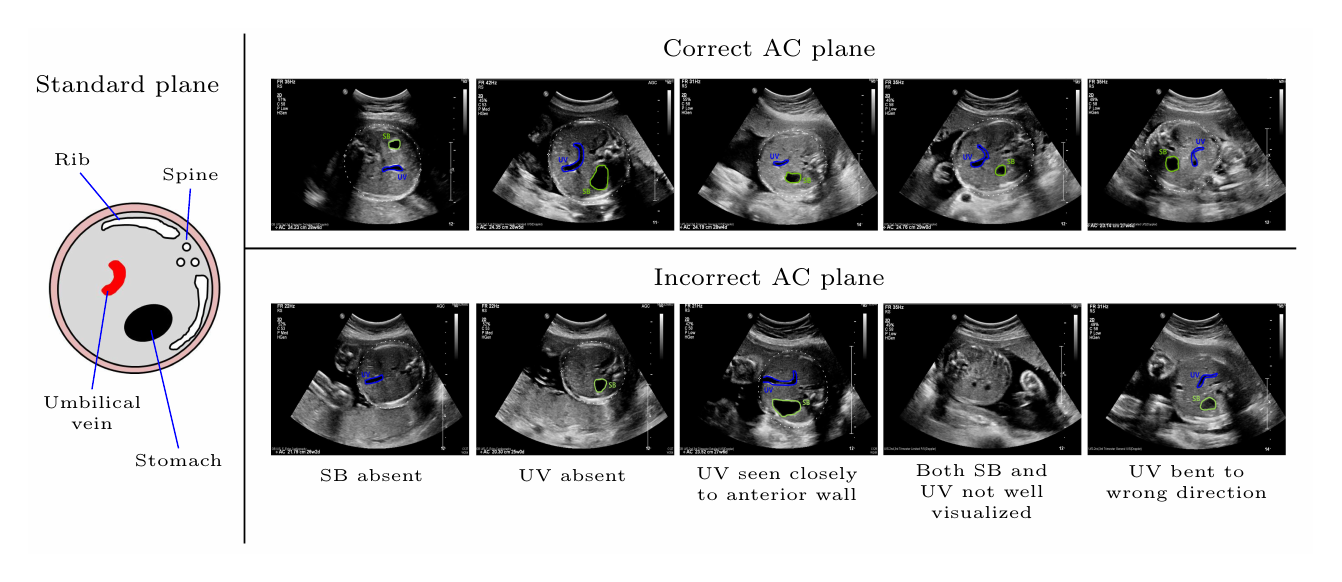}

\caption{Fetal abdominal ultrasound images and anatomical structures. In a standard plane, stomach bubble (SB) and (UV)  appearing as "hockey-stick" bending against SB must be demonstrated. \label{fig:fetalAbdomen}}
\end{figure*}

\begin{figure*}[h!]
\centering
\begin{tikzpicture}
 \definecolor{boxandarrow}{rgb}{0,0.5,1}
 \coordinate (O) at (-4,2);
 \node at (O) {\includegraphics[width = .275\textwidth]{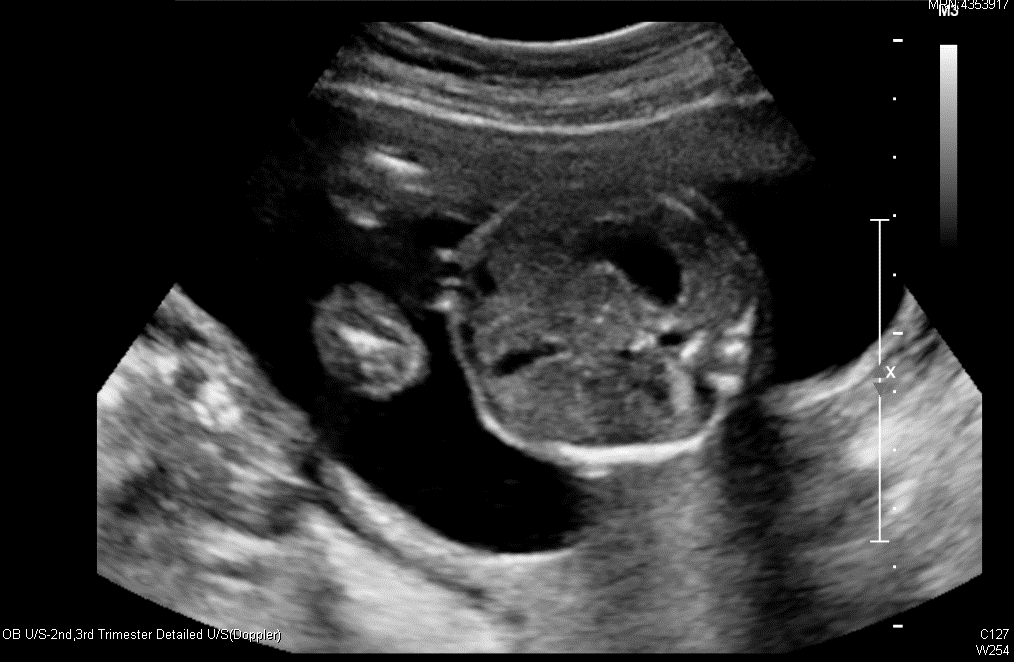}};
 \node at ($(O)-(0,4)$) {\includegraphics[width = .275\textwidth]{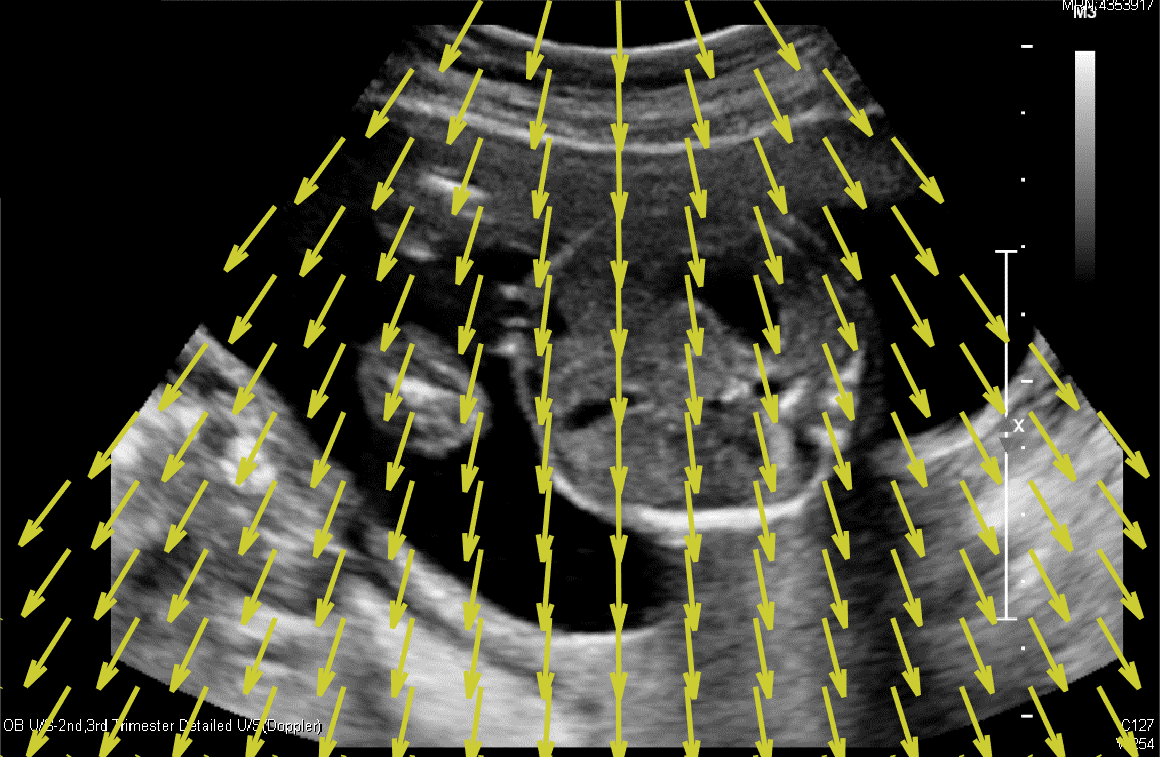}};
 \coordinate (A) at (-3.65,1.7);
 \draw[very thick, color={rgb:red,0;green,176;blue,240}] (A) rectangle ++(1,1);
 \draw[very thick, color=boxandarrow] ($(A)+(.25,.25)$) rectangle ++(.5,.5);
 \draw[very thick, color=boxandarrow] ($(A)-(0,4)$) rectangle ++(1,1);
 \draw[very thick, color=boxandarrow] ($(A)+(1,0.5)$) -- ++(2,0) node[above,color=black,text width = 2.2cm, xshift = .3cm, align = center] {Normal-view Wide-view};
 \draw[->, very thick, color=boxandarrow] ($(A)-(0,4)+(1,0.5)$) -- ++(2,0) --node[midway, right, black, text width = 1.9cm]{Ultrasound propagation direction} ++(0,4) -- ++ (1.5,0);
 \draw[very thick, color=boxandarrow] ($(A)+(4.75,0)$) rectangle ++(1,1) node[midway]{\includegraphics[width=.9cm,height=.9cm]{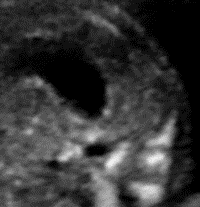}}
 node[midway,yshift=-.8cm,black] {W-view};
 \draw[very thick, color=boxandarrow] ($(A)+(6.25,0)$) rectangle ++(1,1) node[midway]{\includegraphics[width=.9cm,height=.9cm]{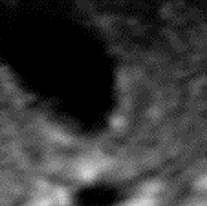}}
 node[midway,yshift=-.8cm,black] {N-view};
 \draw[very thick, color=boxandarrow] ($(A)+(7.75,0)$) rectangle ++(1,1)
 node[midway,yshift=-.8cm,black] {U-Dir};
 \draw[very thick, color=yellow,->]($(A)+(8.1,.8)$) -- ++(.2,-.6);
 \draw[very thick,->, color=boxandarrow] (3,1.2) -- ++(0,-1.5) node[midway,right,color=black]{Input} node[below,color=black]{CNN};
 \coordinate(B) at (1,-1);
 \draw[very thick, color=boxandarrow] (B) rectangle ++(4,-5.5);
  \draw[very thick, color=white] ($(A)+(4.875,-4)$) rectangle ++(1,1) node[midway]{\includegraphics[width=.9cm,height=.9cm]{overall_large.png}};
 \draw[very thick, color=white] ($(A)+(6.125,-4)$) rectangle ++(1,1) node[midway]{\includegraphics[width=.9cm,height=.9cm]{overall_small.png}};
 \draw[very thick, color=white] ($(A)+(7.375,-4)$) rectangle ++(1,1);
 \draw[very thick, color=yellow,->]($(A)+(7.7,-3.2)$) -- ++(.2,-.6);
 \node at ($(B)+(2,-3.2)$) {\includegraphics[width=3.8cm]{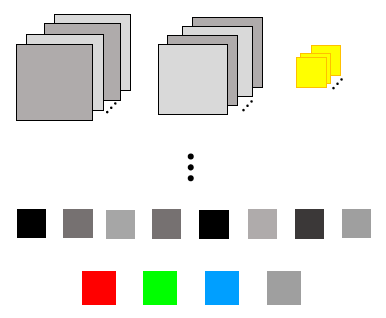}};
 \draw[very thick,->, color=boxandarrow] ($(B)-(.2,4.5)$) -- ++(-2.2,0) node[midway,above,text width = 2.1cm,color=black]{Semantic segmantation} node[midway,below,color=black]{Output} ;
 \node at ($(O)-(0,8)$) {\includegraphics[width = .275\textwidth]{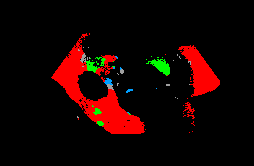}};
 \draw[very thick,->,color = gray] plot [smooth, tension=2] coordinates { ($(B)+(1.75,-4.75)$)  ($(B)+(-1,-5.5)$) ($(A)+(0.5,-7.5)$)};
 \draw[very thick,->,color=boxandarrow]($(O)-(0,9.7)$) -- ++(0,-.6) node [below,color=black] {AC fitting};
 \draw[very thick,->,color=boxandarrow]($(O)+(2.75,-12.7)$) -- ++(2,0) -- ++(0,3) -- ++(2.25,0)  --++(0,-.6) node [below,color=black] {Plane acceptance check};
 \node at ($(O)-(0,12.5)$) {\includegraphics[width = .275\textwidth]{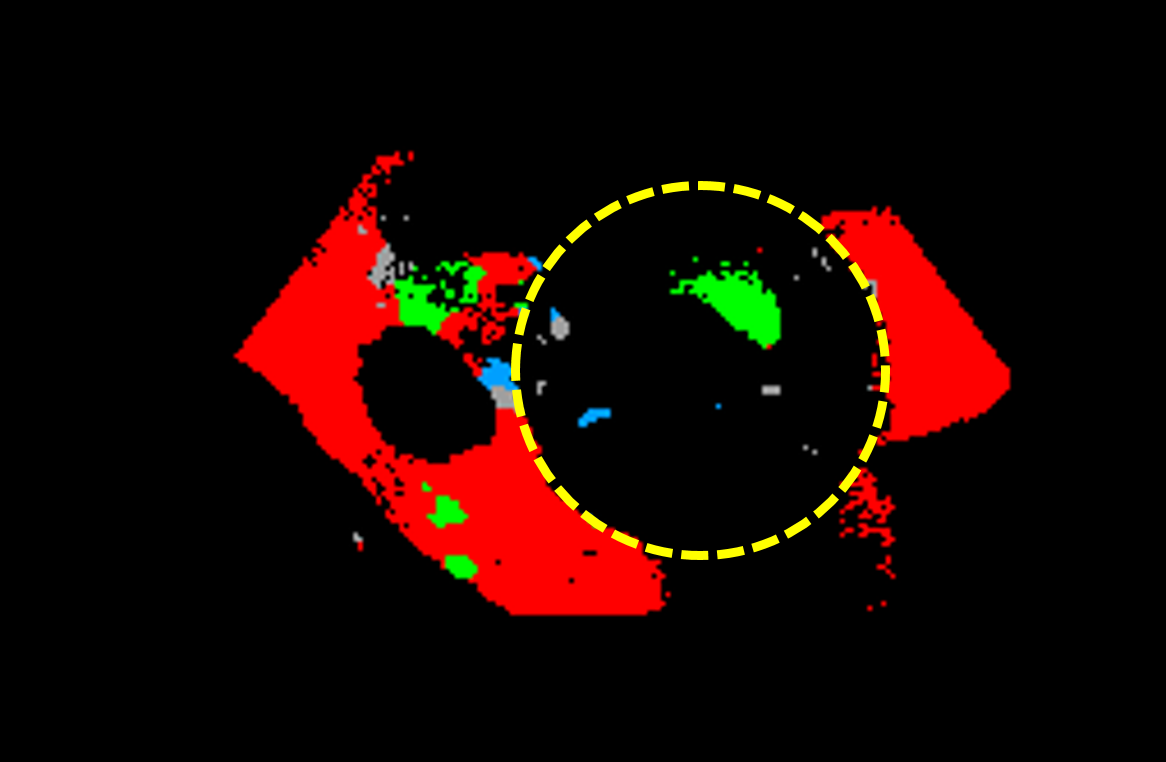}};
 \node at ($(O)-(-7,12.5)$) {\includegraphics[width = .2\textwidth]{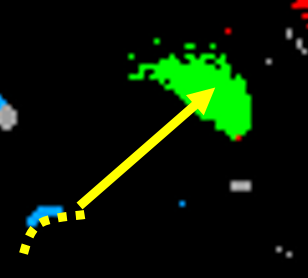}};
 \draw[thick,->]($(B)+(2,-1.375)$) -- ++(0,-.25);
 \draw[thick,->]($(B)+(2,-2.8)$) -- ++(0,-.25);
 \draw[thick,->]($(B)+(2,-4.05)$) -- ++(0,-.25);
 \node at ($(B)+(1.05,-5)$) {AF};
 \node at ($(B)+(1.7,-5)$) {SB};
 \node at ($(B)+(2.35,-5)$) {UV};
 \node at ($(B)+(3,-5)$) {SA};
\end{tikzpicture}
\caption{ {Overall process of the proposed framework.} The proposed framework performs semantic segmentation by using a CNN, AC measurement, and plane acceptance check. Especially, the CNN used for the semantic segmentation admits {\it normal-view} (N-view), {\it wide-view} (W-view), and ultrasound propagation direction (U-Dir). \label{fig:overallProcess}}
\end{figure*}

Fetal AC measurement requires a suitable selection of transabdominal ultrasound images, as shown in Fig. \ref{fig:fetalAbdomen}, and the identification of the fetal region from the noisy ultrasound images.
The standard AC plane must contain the stomach bubble(SB) and the portal section from the umbilical vein(UV), which has the characteristic ``hockey-stick'' appearance \cite{Salomon2010}.
Additionally, there exists a portion of the fetal boundary that overlaps with a portion of the amniotic fluid (AF) boundary.
To utilize these facts, SB, UV, and AF should be distinguished from each other and shadowing artifacts (SA) because all of them appear as anechoic region in B-mode ultrasound images.

Taking account of these observations, the proposed method consists of three main steps: anatomical structure detection using CNN, fetal abdominal region detection using Hough transform, and acceptance plane verification using another CNN (Fig. \ref{fig:overallProcess}).
Before explaining the proposed method in detail, let us review the CNN.

\subsection{Proposed CNN Structure\label{sec:CNNbasic}}
{ 
CNN is a type of artificial neural network inspired by visual information processing in the brain.
To recognize complex features from the visual information, CNN consists of several layers, which extract and repeatedly combine low-level features for composing high-level features.
The composed high-level features are used to classify an input image.

Generally, many CNNs consist of combinations of convolutional, pooling, and fully connected layers.
The convolutional layer (C-layer) extracts higher-level features by convolving received feature maps from the previous layer and activating the convolved features.
In this paper, the rectified linear unit (ReLU) $g(x) = max(0,x)$ is used as the activation function.
A C-layer is usually followed by a pooling layer (P-layer), which reduces the dimensions of feature maps by ``max pooling.''
The max pooling downsamples the input feature maps by striding a rectangular receptive field and taking the maximum in the field.
After C-layers and P-layers, a fully connected layer (F-layer) integrates high-level features and produces compact feature vectors.
Like the C-layers, ReLU is used as the activation function of the F-layers in our research.
On the final layer, say the $J$-th layer, the output layer produces the posterior probability $\mathbf{p}$ for each class.
Classification is achieved by finding label corresponding to the maximum of $\mathbf{p}$.

Training a CNN desires to find proper parameters of the CNN, say $\theta$, using training data.
To find a proper $\theta$, entropy or energy $L(\mathbf{x}_k,\mathbf{y}_k;\theta)$ is defined and desired to be minimized where $(\mathbf{x}_k, \mathbf{y}_k)$ denotes training data.
In other words, a proper set of parameters $\hat\theta$ is obtained by the optimization problem
\begin{equation}
\hat\theta = \arg\min_{{\theta}}\sum_{k} L(\mathbf{x}_k,\mathbf{y}_k;{\theta}).
\end{equation}
For more details about CNN, readers may refer to \cite{LeCun1989, LeCun2015}.
}

\begin{figure}
\centering
 \includegraphics[width = 0.48\textwidth]{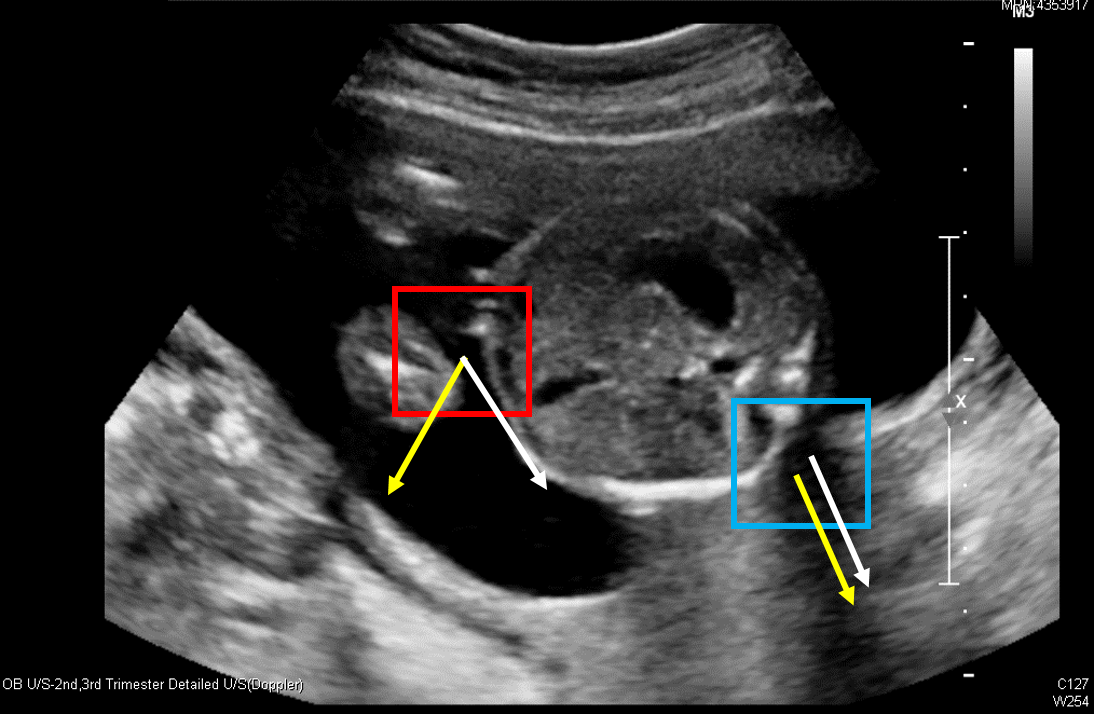}
  \caption{ {The relation between ultrasound propagation direction (white arrows) and image pattern direction (yellow arrows) in image patches.} In the image patch corresponding to shadowing artifact (light blue box), the image pattern is strongly related to the ultrasound propagation direction compared to the patch corresponding to AF (red box).\label{fig:observationFromImages}}
\end{figure}
\begin{figure}
\centering
 \includegraphics[width = 0.48\textwidth]{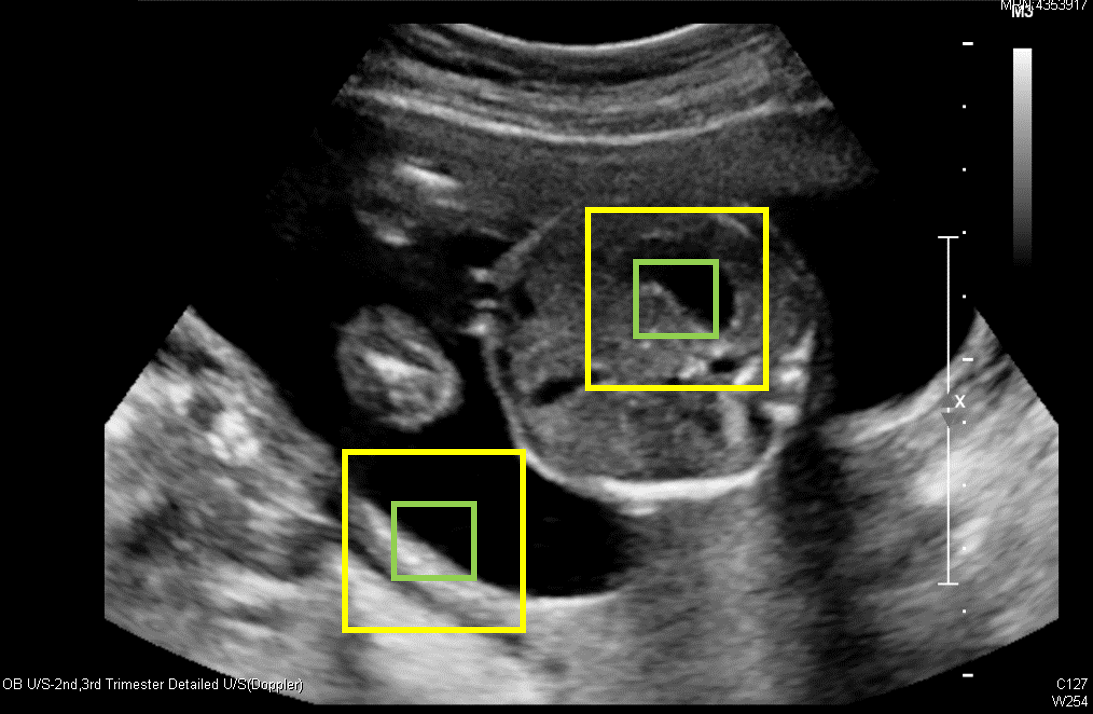}
  \caption{ {Variation of observed image feature according to the size of a view.} Local pattern of dark region appear similar in a relatively small size of view (two green boxes).
However, as the size of view increases (two yellow boxes), distinct image feature appears in each view.\label{fig:observationFromImages2}}
\end{figure}
The proposed CNN structure is based on the following observations, motivated by the comparison between the sonographers' classification process and conventional CNN structures for object recognition:
\begin{enumerate}
 \item A position is classified as a shadowing artifact not only by the local image pattern at the position but also by the expected ultrasound propagation direction and the position of hard materials (spine, ribs) (Fig. \ref{fig:observationFromImages}).
 \item For accurate classification, both the local and non-local structure information need to be integrated.
     For example, in Fig. \ref{fig:observationFromImages2}, we can observe two positions having similar local patterns with distinct non-local structures.
\end{enumerate}
Based on the first observation, the proposed CNN structure admits the ultrasound propagation direction as one of input.
For example, the propagation direction $(u,v)$ can be simply modelled by
\begin{equation}
 u = \f{x}{\sqrt{x^2+y^2}}, v = \f{y}{\sqrt{x^2+y^2}}
\end{equation}
where $(x,y)$ is the position with respect to the probe position.
For the computational efficiency of the CNN, the size of input should be as small as possible.
However, as per our second observation, we need both the local and non-local structure information to classify each position accurately.
Therefore, we used two image patches corresponding to a normal view and a wide view as inputs.
Firstly, a $128\times128$ sized local image patch was selected as the {\it normal-view} image patch to analyze the local structure around a given position by using the image pattern.
Secondly, a $256\times256$ sized non-local image patch was selected as the {\it wide-view} image patch to combine the local structure with other structures far from the position.
To reduce computational cost, the wide-view image patch was simplified into a low-resolution image of size $128\times128$.

The output of the proposed CNN structure is a $1\times1\times4$ vector, which corresponds to 4 categories of SA and the main anatomical structures in the standard abdominal plane: SB, UV, and AF.
We chose all the image patches centered at dark pixels ($\leq 0.1\times\text{Max. intensity}$) in a given ultrasound image.
Then, the proposed CNN allows the classification of the chosen image patches into 4 categories.
With this result, a semantically segmented image is made by coloring each of the chosen pixels according to their categories.

The proposed CNN begins with three branches, which are designed to handle the propagation direction and each image patch of multiple views.
Each branch extracts the desired image features and analyzes the propagation direction.

{\footnotesize
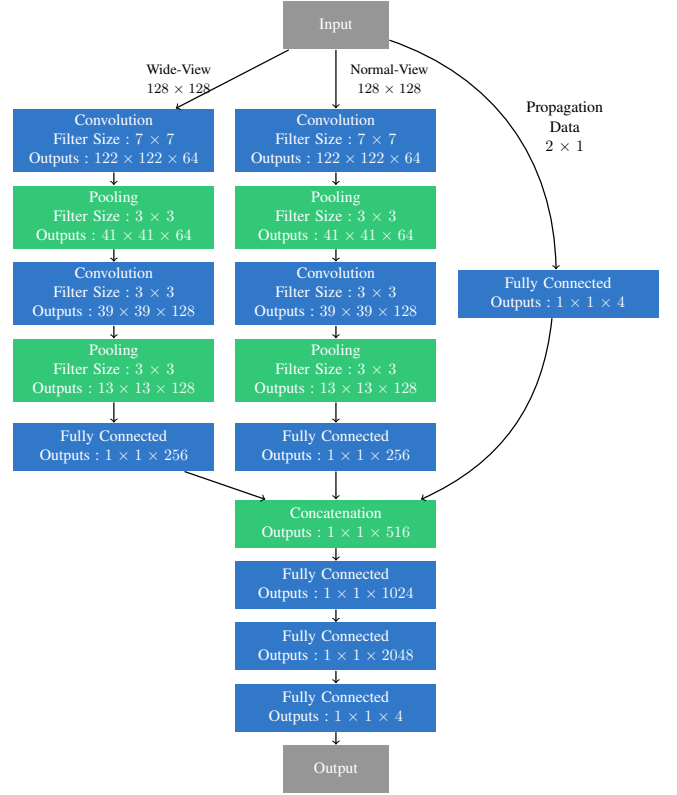
\begin{figure}
\scalebox{.6}{
  \begin{tikzpicture}
\node [input] (input) at (0,0){Input};

\node [C] (LC1) at (\largeImageXPosition,\layerDist*1.5){Convolution\\Filter Size : $7 \times 7$\\Outputs : $122\times 122 \times 64$};
\node [C] (SC1) at (\smallImageXPosition,\layerDist*1.5){Convolution\\Filter Size : $7 \times 7$\\Outputs : $122\times 122 \times 64$};
\node [P] (LP1) at (\largeImageXPosition,\layerDist*2.5){Pooling\\Filter Size : $3 \times 3$\\Outputs : $41\times 41 \times 64$};
\node [P] (SP1) at (\smallImageXPosition,\layerDist*2.5){Pooling\\Filter Size : $3 \times 3$\\Outputs : $41\times 41 \times 64$};
\node [C] (LC2) at (\largeImageXPosition,\layerDist*3.5){Convolution\\Filter Size : $3 \times 3$\\Outputs : $39\times 39 \times 128$};
\node [C] (SC2) at (\smallImageXPosition,\layerDist*3.5){Convolution\\Filter Size : $3 \times 3$\\Outputs : $39\times 39 \times 128$};
\node [P] (LP2) at (\largeImageXPosition,\layerDist*4.5){Pooling\\Filter Size : $3 \times 3$\\Outputs : $13\times 13 \times 128$};
\node [P] (SP2) at (\smallImageXPosition,\layerDist*4.5){Pooling\\Filter Size : $3 \times 3$\\Outputs : $13\times 13 \times 128$};
\node [F] (LF) at (\largeImageXPosition,\layerDist*5.5){Fully Connected\\Outputs : $1\times 1 \times 256$};
\node [F] (SF) at (\smallImageXPosition,\layerDist*5.5){Fully Connected\\Outputs : $1\times 1 \times 256$};

\node [F] (PF) at (\propagationXPosition,\layerDist*3.5){Fully Connected\\Outputs : $1\times 1 \times 4$};

\node [conc] (Con) at (0,\layerDist*6.5){Concatenation\\Outputs : $1\times 1 \times 516$};

\node [F] (TF1) at (0,\layerDist*7.3){Fully Connected\\Outputs : $1\times 1 \times 1024$};
\node [F] (TF2) at (0,\layerDist*8.1){Fully Connected\\Outputs : $1\times 1 \times 2048$};
\node [F] (TF3) at (0,\layerDist*8.9){Fully Connected\\Outputs : $1\times 1 \times 4$};

\node [O] (out) at (0,\layerDist*9.7){Output};

\path [draw, ->, thick] (input) -- (LC1) node [midway, right, text width=6em, text centered] (TextNode1) {\small{Normal-View\\ $128 \times 128$}};
\path [draw, ->, thick] (LC1) -- (LP1);
\path [draw, ->, thick] (LP1) -- (LC2);
\path [draw, ->, thick] (LC2) -- (LP2);
\path [draw, ->, thick] (LP2) -- (LF);
\path [draw, ->, thick] (LF) -- (Con);

\path [draw, ->, thick] (input) -- (SC1) node [midway, left, text width=6em, text centered] (TextNode1) {\small{Wide-View\\ $128 \times 128$}};
\path [draw, ->, thick] (SC1) -- (SP1);
\path [draw, ->, thick] (SP1) -- (SC2);
\path [draw, ->, thick] (SC2) -- (SP2);
\path [draw, ->, thick] (SP2) -- (SF);
\path [draw, ->, thick] (SF) -- (Con);

\path [draw, ->, thick] (input) edge[bend left=35] node[midway,right, text width=6em, text centered] {Propagation Data \\ $2\times 1$} (PF);
\path [draw, ->, thick] (PF) edge[bend left] (Con);

\path [draw, ->, thick] (Con) -- (TF1);
\path [draw, ->, thick] (TF1) -- (TF2);
\path [draw, ->, thick] (TF2) -- (TF3);
\path [draw, ->, thick] (TF3) -- (out);

\end{tikzpicture}
}
 \caption{ {The proposed CNN structure.} The CNN uses input data as a combination of image patches of multiple sizes and ultrasound propagation direction. From the image patches and the propagation directions, feature vectors are extracted and combined to classify a given image patch. \label{fig:CNNStructure}}
\end{figure}
}
As shown in Fig. \ref{fig:CNNStructure}, two branches for image analysis basically consist of pairs of convolutional and max-pooling layers as well as a fully connected layer. In the branch analyzing the {\it normal-view} image patch, the first and second convolutional layers, respectively, used $7\times7$ and $3\times3$ filters, and the $3\times3$ max-pooling was used with a stride step of $2$ in both max-pooling layers.
The branch analyzing the {\it wide-view} image patch consists of the same structure as the combinations of the convolutional and max-pooling layers, too.
The ultrasound propagation direction is analyzed through a fully connected layer to detect the propagation direction.

The result produced by each branch is concatenated into one feature vector for classification. This vector passes through two fully connected layers to classify the given data into 4 classes. We made a semantic segmentation image using this classification results (Fig. \ref{fig:overallProcess}).

\subsection{Measurement Agreement\label{sec:hough}}
\begin{figure*}[h!]

\centering
\begin{tikzpicture}[xscale = .3, yscale = .3]
\coordinate (a) at (0,12);
\node at ($(a)$) {\small (a)};
\node at ($(a)+(0,4.5)$) {\includegraphics[width=2.5cm]{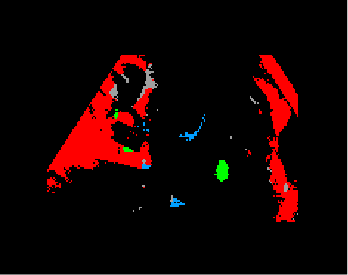}};
\node at ($(a)+(9,0)$) {\small (b)};
\node at ($(a)+(9,4.5)$) {\includegraphics[width=2.5cm]{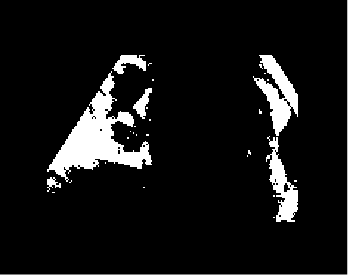}};
\node at ($(a)+(18,0)$) {\small (c)};
\node at ($(a)+(18,4.5)$) {\includegraphics[width=2.5cm]{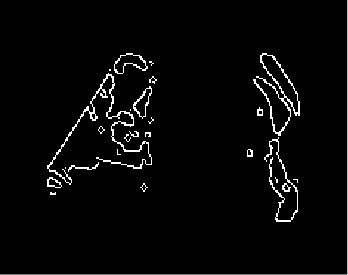}};
\node at ($(a)+(9,-9)$) {\small (d)};

\node at ($(a)+(0,-4.5)$) {\includegraphics[width=2.5cm]{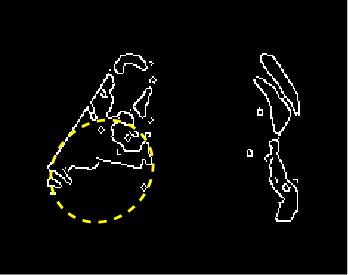}};
\node at ($(a)+(9,-4.5)$) {\includegraphics[width=2.5cm]{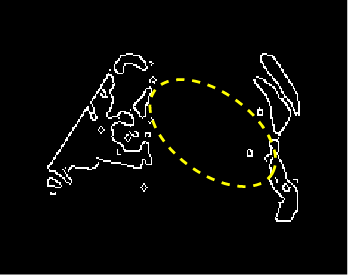}};
\node at ($(a)+(18,-4.5)$) {\includegraphics[width=2.5cm]{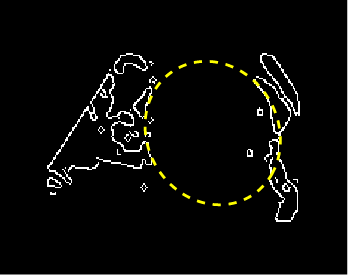}};

\node at ($(a)+(30,-9)$) {\small(e)};
\node at ($(a)+(30,0)$) {\includegraphics[width=4cm,height=2.83cm]{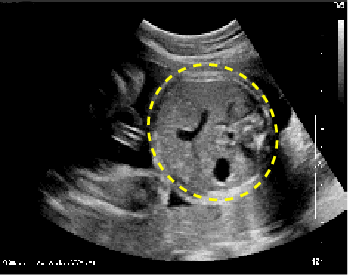}};

\node at ($(a)+(44.5,0)$) {\includegraphics[width=4cm]{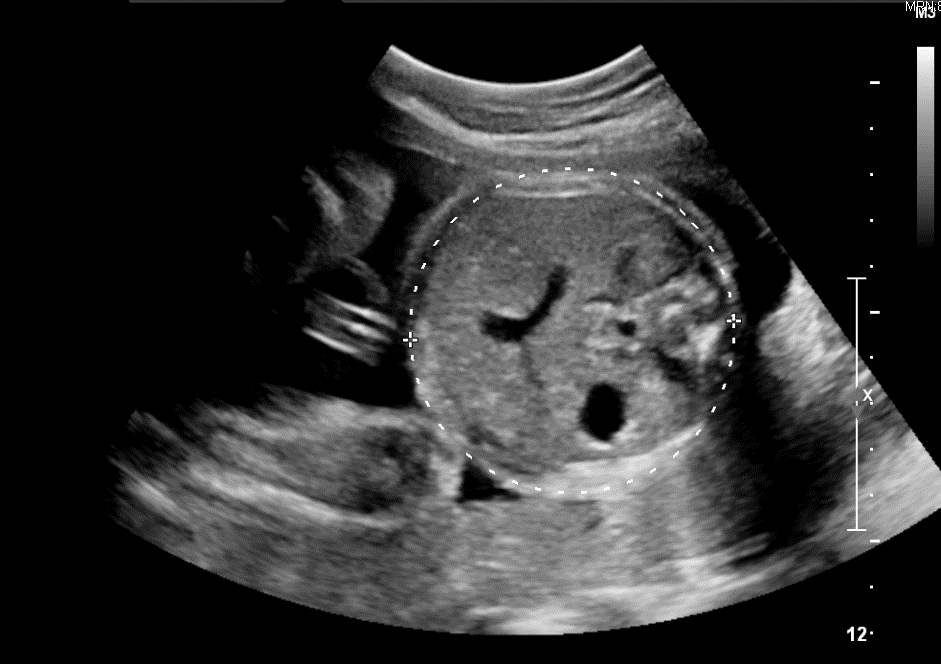}};
\node at ($(a)+(44.5,-9)$) {\small (f)};

\end{tikzpicture}

\caption{ {The fetal abdomen detection work flow.} (a) is acquired sementic segmentation image. (b) and (c) are extracted AF region and each boundary images respectively. (d) are candidates of fetal abdomen generated by Hough transform. With these candidates, the best fitting ellipse which we choose is shown in (e) with an experts' caliper placement (f).\label{fig:ellipseresult}}
\end{figure*}

The commonly used AC measurement is the manual fitting of an ellipse (or circle) to a fetal abdominal contour. In order to detect this ellipse form automatically, the ellipse detection method based on Hough transform \cite{YX2002,MC1998,Xu1990,BN1999} has been proposed. However, direct application of these methods to extracted AF region could produce undesired ellipse candidates (Fig. \ref{fig:ellipseresult}(d)), since the AF region in our semantic image does not surround the entire fetal abdomen region (Fig. \ref{fig:ellipseresult}(c)).
To select the proper ellipse out of candidates generated from the ellipse detection method \cite{YX2002}, we only accept ellipses with the ratio of minor axis to major axis greater than 0.6. Among remaining candidates, the half of the candidates which have less amount of AF are selected as our fetal abdominal boundary (Fig. \ref{fig:ellipseresult}(e)).
For a stable result, the medians of major axis, minor axis, center, and angle from the positive horizontal axis to the major axis of the candidate ellipses are taken as parameters of a final ellipse.
We estimate AC by calculating the selected ellipse boundary using
\begin{equation}\label{eq:AC}
  \text{AC}\approx\pi\left[3(a+b)-\sqrt{(3a+b)(a+3b)}\right],
\end{equation}
where $a$ and $b$ are the major and minor axis of ellipse.

Transabdominal images demonstrating proper landmarks for the true axial plane for AC measurement were obtained by an expert.
The AC measurement was performed either manually by other experts or by using the proposed method.
The assessment of localization of fetal abdomen region and the comparison of AC values between the manual and CNN method was performed.

\subsection{Plane Acceptance Check\label{sec:planeAcceptance}}
\begin{figure}[h!]
  \begin{tikzpicture}
 \node (Input) at (0, 0) {\includegraphics[height=2.5cm]{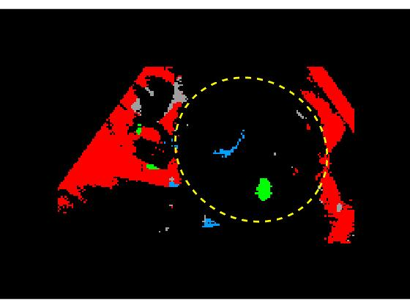}};
 \node at ($(Input)+(0,-1.5)$) {\small(a)}; 
 \node (ROI) at (3.7, 0) {\includegraphics[height=2.5cm]{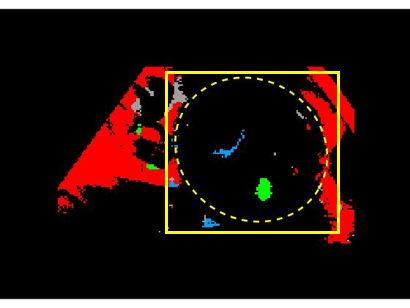}};
 \node at ($(ROI)+(0,-1.5)$) {\small(b)}; 
 \node (Cut) at (6.2, 0) {\includegraphics[height=1cm,width=1cm]{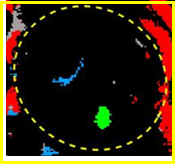}};
 \node at ($(Cut)+(0,-1.5)$) {\small(c)}; 
\end{tikzpicture}
\caption{The process for the acceptance check. (a) is an semantic segmentation image with its detected ellipse. Based on the detected ellipse, semantic segmentation image in a abdominal region is cropped like the yellow box in (b), and (c) the cropped image is rescaled image and used as an input of the CNN for the acceptance check. \label{fig:accepWF}}
\end{figure}

In this section, we evaluate the suitability of the selected plane to determine whether the plane is appropriate for measuring AC.
The semantic segmentation image is cropped to the estimated fetal abdomen area from the previous step and the cropped image is rescaled to be $128\times128$ size as described in Fig. \ref{fig:accepWF}.
Especially, gray region corresponding to the shadowing artifact is excluded when the image is rescaled.
By admitting the rescaled image as the input, CNN in Table \label{tbl:CNNstructure2} estimates the suitability with the probability that the given image is appropriate for measuring AC.
The CNN consists of three pairs of convolutional and pooling layers, and three fully-connected layers.
The first convolutional layer detects features from different channels (RGB) which mean different anatomical structures, and the feature information is propagated through the following convolutional layers to analyse anatomical configuration.
Last three fully-connected layers integrate the detected features and determine the suitability.

Transabdominal ultrasound images were obtained by an expert and reviewed by each of the two ultrasound experts, including the operator.
When either of the experts accepts a given image, the given image was considered as an acceptable image.
The proposed CNN were evaluated by comparing its acceptance result to experts' acceptance result.

\begin{table}[h!]
 \caption{The proposed CNN structure for plane acceptance check. The output of the network has 8 classes which correspond to the 8 directions. \label{tbl:CNNStructure2}}
\centering
 \begin{tabular}{c | c c c c}
 \hline
 \multirow{2}{*}{Input}
 	& \multicolumn{4}{c}{Umbilical vein image} \\\cline{2-5}
    & \multicolumn{4}{c}{$128\times128\times3$} \\	
 \hline
  	& Type & Maps & Filter size & Stride \\
   							& C &  $124\times124\times64$ & $5\times5\times3$  & - \\
   							& P & $41\times41\times64$ & $3\times3$ & 2	  \\
   							& C & $39\times39\times128$ & $3\times3$  & -\\
   							& P &$20\times20\times128$  & $2\times2$  & 2\\
   							& F & $1\times1\times256$ & - & -  \\
   							& F & $1\times1\times512$ & - & -  \\
   							\hline
   		Output	& F & $1\times1\times2$ & -  \\
  \hline
   	
 \end{tabular}
\end{table}

\section{Results}
\subsection{Data and Experimental Setting}
For training and evaluation, fetal abdominal ultrasound images were provided by the department of Obstetrics and Gynecology, Yonsei university college of medicine, Seoul, Korea (IRB no.: 4-2017-0049).
{ 
We were provided with 88 cases of ultrasound images and each case consists of several true and false abdominal ultrasound images obtained from a pregnant woman by experts with an IU22 (Phillips, Seoul, Korea) ultrasound machine and a 2--6-MHz transabdominal transducer.
The provided cases were separated into ``training cases'' and ``test cases'' which are consist of 56 cases and 32 cases, respectively.
The training images were used to generate training data for the classification CNN and the acceptance check CNN, and tune a heuristic parameters for the ellipse detection.
}

{\it Caffe} \cite{Jia2014} was used to implement and train the two proposed CNN in our framework.
Our framework which consists of the proposed CNNs and the Hough transform-based ellipse detection \cite{Simonovsky2013} was implemented with MATLAB and Python.

\subsection{Training Performance of the proposed CNN}
{  As mentioned, fetal abdominal ultrasound images from the 56 test cases were provided and 13261 pairs of image patches of multiple views were extracted from the images with the ultrasound propagation direction in those patches.
The extracted patches was divided into 2 sets, training set and test set, to evaluate training process by the simple cross validation.
The ratio of the training set to the test set are approximately 2:1. }
We used ADAM to minimize the loss function \cite{Kingma2014} and dropout ratio $0.5$ on the last layer during training to prevent overfitting \cite{Srivastava2014}.

\begin{figure}
\centering
\scalebox{1.2}
{
\begin{tikzpicture}
 \node at (-0.17, -1) {\includegraphics[height=2cm]{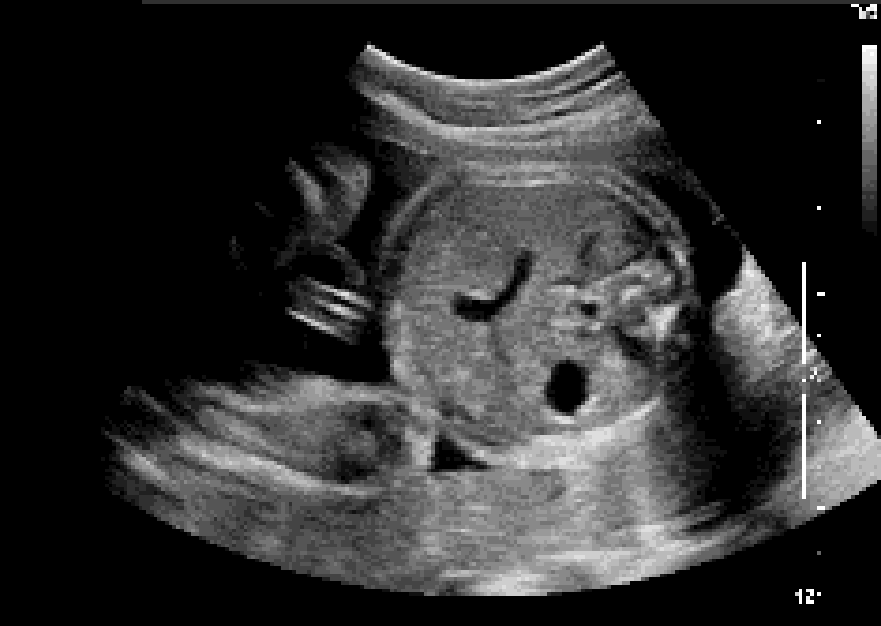}};
 \node at (0, 1.5) {\includegraphics[height=2cm]{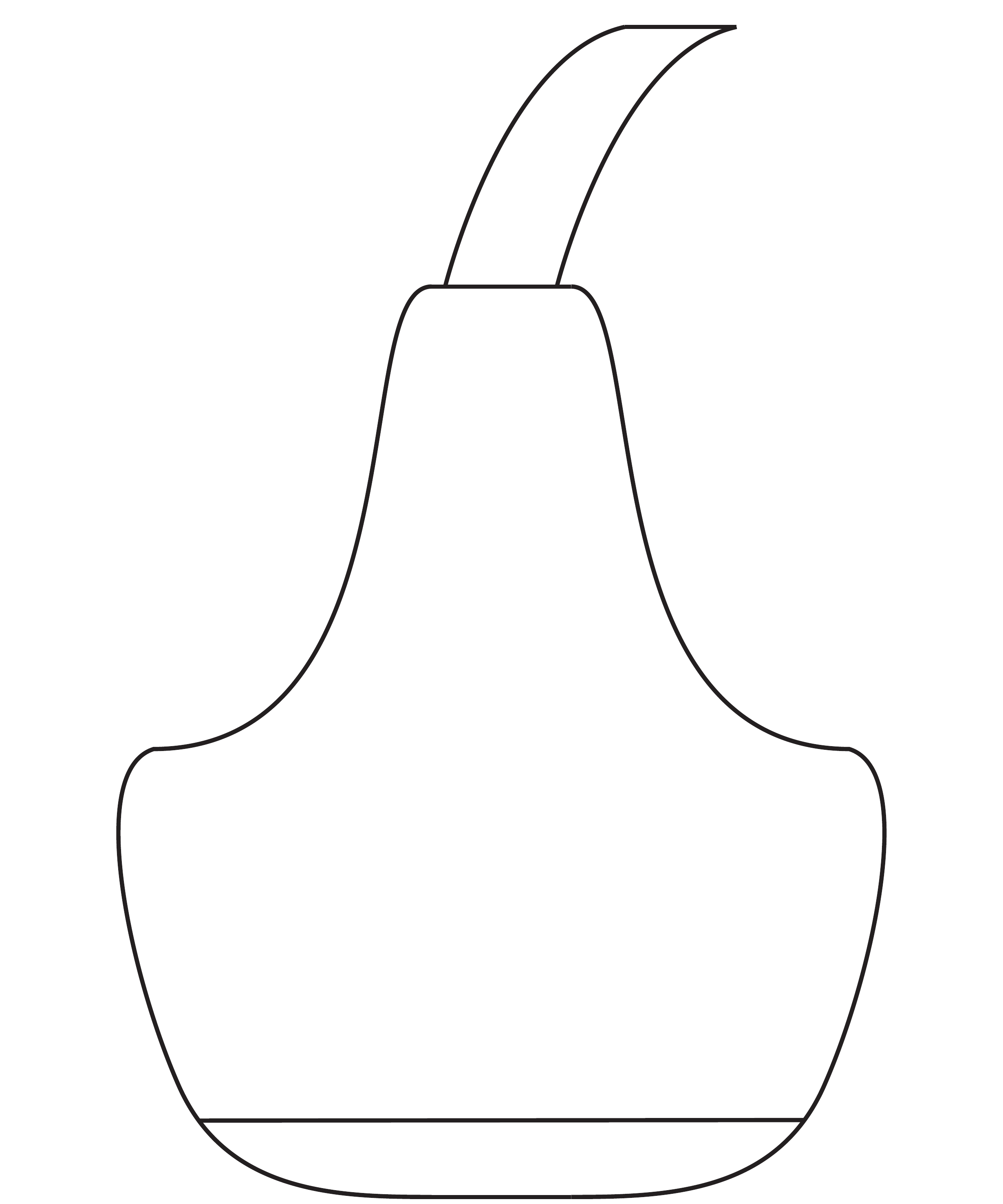}};
 \draw[->, very thick, color = red] (0,.5) -- ++(-1.3466,1.4787);
 \draw[->, very thick, color = red] (0,.5) -- ++(1.9737,-0.3233);
 \draw[->, very thick, color = red] (0,.5) -- ++(0.1385,1.9952);
 \draw[->, very thick, color = red] (0,.5) -- ++(0.8126,-1.8275);
 \node at (0,-2.3) {\scriptsize (a)};

 \node at (3.33, -1) {\includegraphics[height=2cm]{uso.png}};
 \node at (3.5, 1.5) {\includegraphics[height=2cm]{transducer.pdf}};
 \draw[->, very thick, color = red] (3.5,.5) -- ++(1,-1.7321);
 \draw[->, very thick, color = red] (3.5,.5) -- ++(-0.3473,-1.9696);
 \draw[->, very thick, color = red] (3.5,.5) -- ++(-1,-1.7321);
 \draw[->, very thick, color = red] (3.5,.5) -- ++(0.3473,-1.9696);
 \node at (3.5,-2.3) {\scriptsize (b)};
\end{tikzpicture}
}
\caption{The initialized filters of the fully-connected layer for propagation direction. Filters are initialized so that each component of filters follows Gaussian distribution (a) and the filters are uniformly distributed toward the imaging range (b). (f) is an expert's example of AC caliper placement. Compared to automatic caliper placement, the expert's placement tends to be smaller. \label{fig:dirFilter}}
\end{figure}

\begin{figure*}
\centering
\subfigure[]
{\includegraphics[width=.4\textwidth]{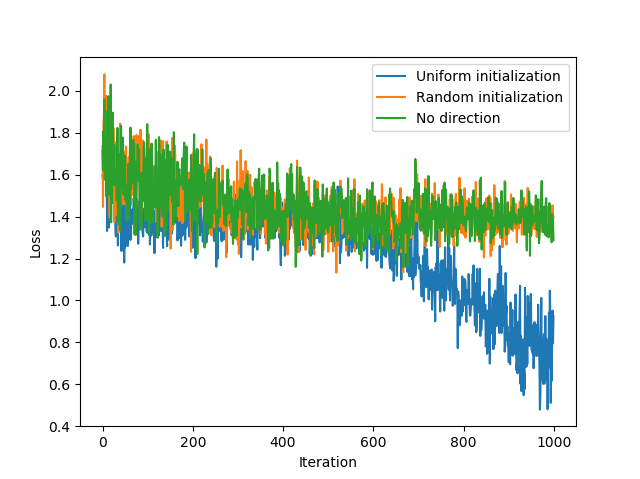}}
\subfigure[]
{\includegraphics[width=.4\textwidth]{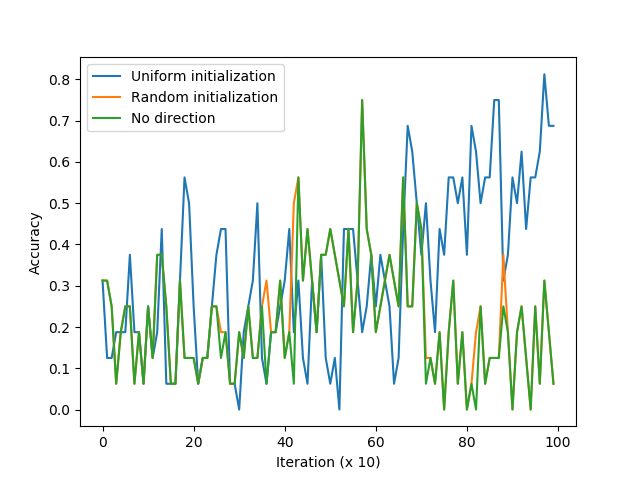}}
\caption{Train loss (a) and test accuracy (b). We trained the following three casese : 1. the filters are initialized to be unifomrly distributed toward the imaging domain (blue lines). 2. the filters are initialized randomly with the Gaussian distribution (green lines). 3. ultrasound propagation direction is not used for the classification (red lines).
On the other hand, in the image branches, the weights were initialized with the same values for the three cases. \label{fig:learningCurve}}
\end{figure*}

Vectors in Fig. \ref{fig:dirFilter} correspond to initialized filters, say directional filters, of the fully-connected layer in a branch for propagation direction analysis, say direction branch.
Fig. \ref{fig:dirFilter} shows the directional filters initialized as randomly distributed vector with a normal distribution and as uniformly distributed vectors.
Training processes with the two initialization strategies were compared to the training process without the direction analysis branch.
In this comparison, filters in image branches were initialized with same filters for the three cases.
The training loss and test accuracy changes are plotted as graph in Fig. \ref{fig:learningCurve}(a) and (b), respectively.
When the directional filters are randomly initialized , not all filter vectors could be toward the image range and some vectors are obtuse for all ultrasound propagation direction in the image range as described in Fig. \ref{fig:dirFilter}(b).
Because the inner-products between the obtuse vectors and ultrasound propagation directions are negative, neurons corresponding to the obtuse filter vectors is not activated by ReLU function and the filters are not updated during training.
In Fig. \ref{fig:learningCurve}, the difference of training performances is not notable between the cases with randomly initialized directional filter and without the direction analysis branch.
On the other hand, when the directional filters are initialized to be uniformly distributed toward the image range, all directional filters contribute classification.
Therefore, it results that convergence speed increases in the uniform initialization case.
In the following sections, we use the trained filters with the initialized directional filters as uniformly distributed vector toward the imaging range.

\subsection{AC Measurement}
\begin{figure*}
\centering
\subfigure[]
{
\includegraphics[width = .23\textwidth]{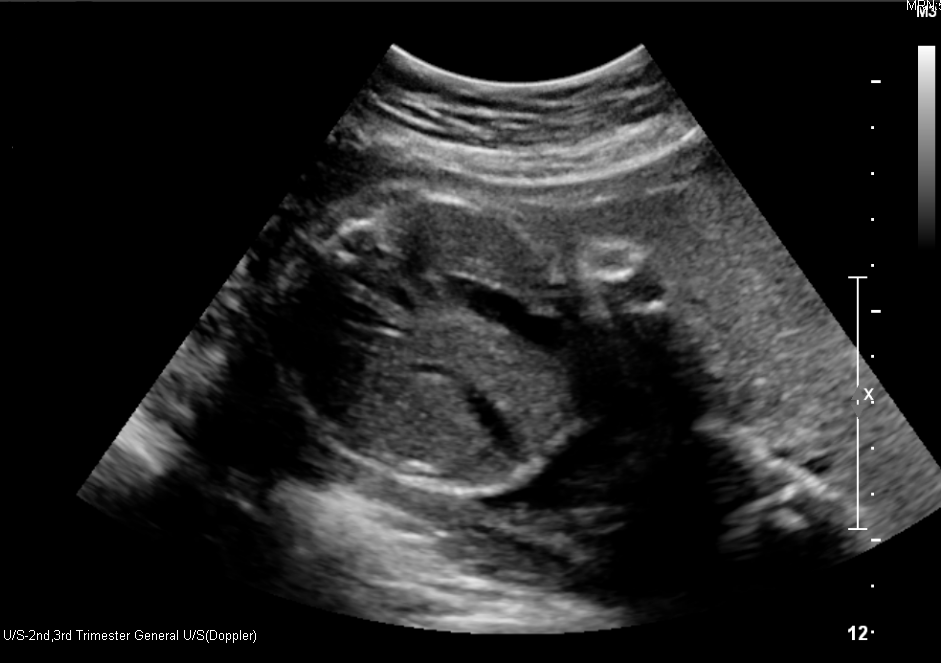}}
\subfigure[]
{
\includegraphics[width = .23\textwidth]{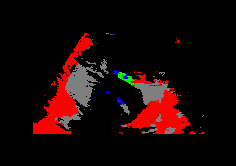}}
\subfigure[]
{
\includegraphics[width = .23\textwidth]{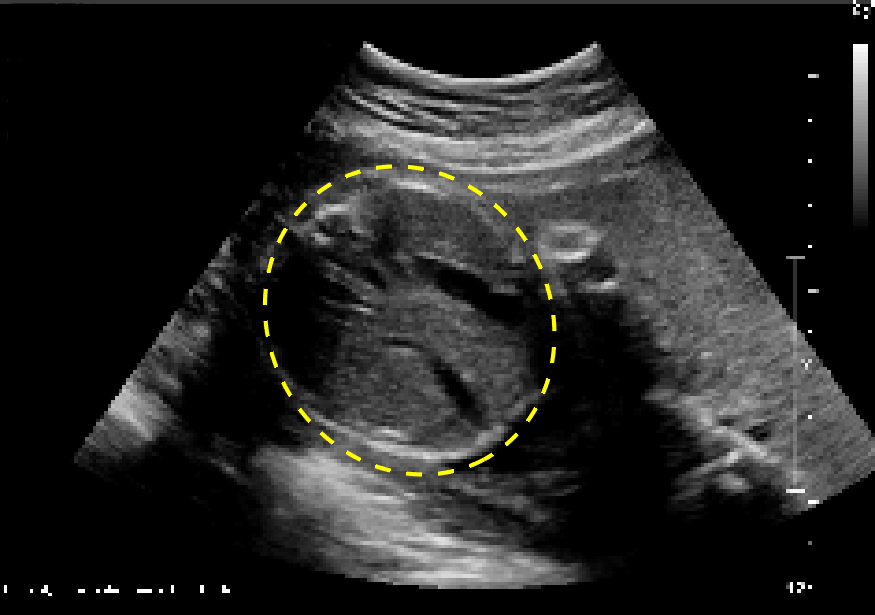}}
\subfigure[]
{
\includegraphics[width = .23\textwidth]{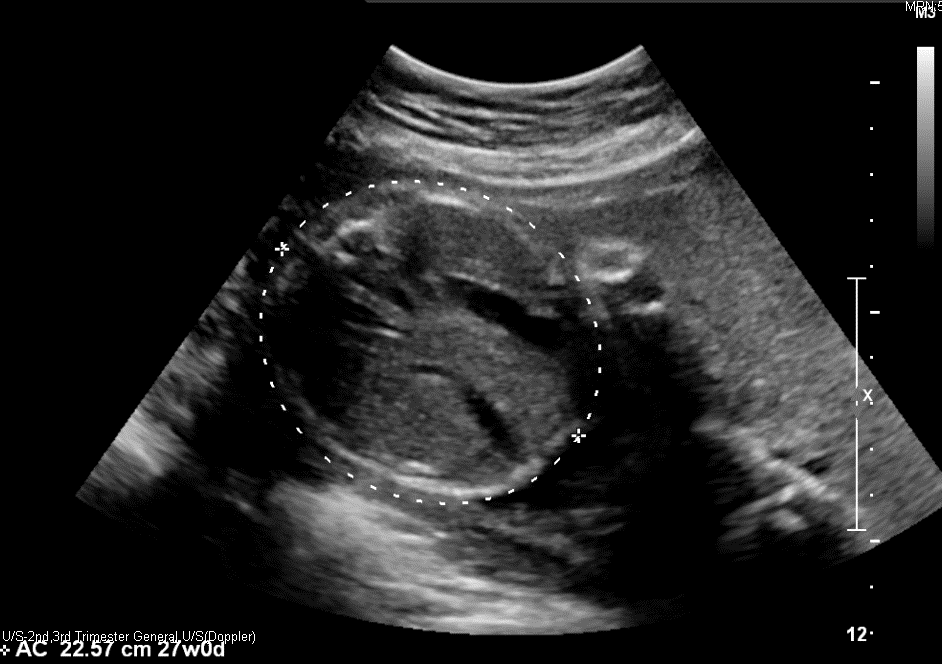}}

\subfigure[]
{
\includegraphics[width = .23\textwidth]{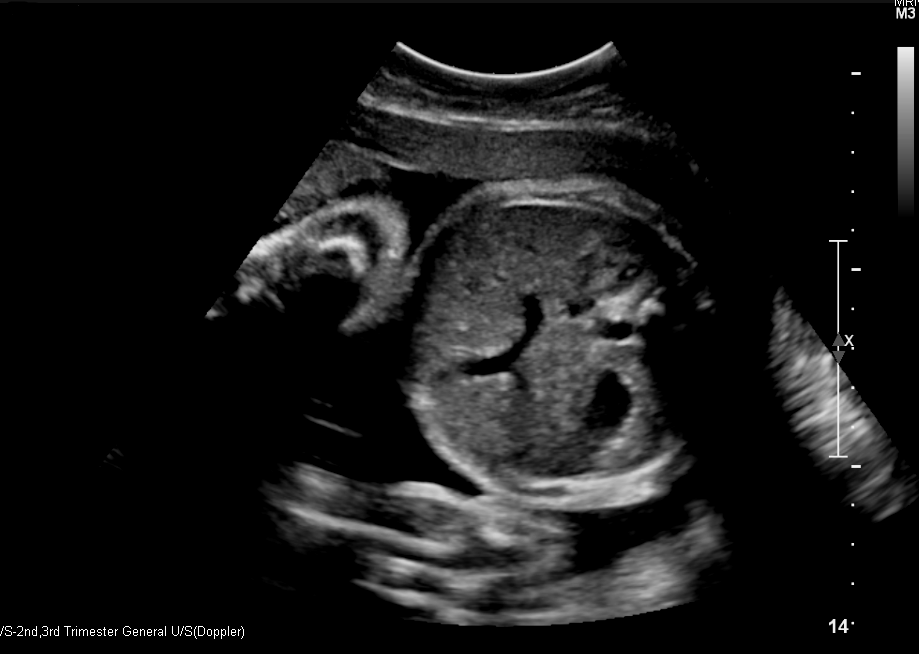}}
\subfigure[]
{
\includegraphics[width = .23\textwidth]{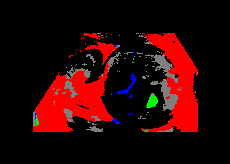}}
\subfigure[]
{
\includegraphics[width = .23\textwidth]{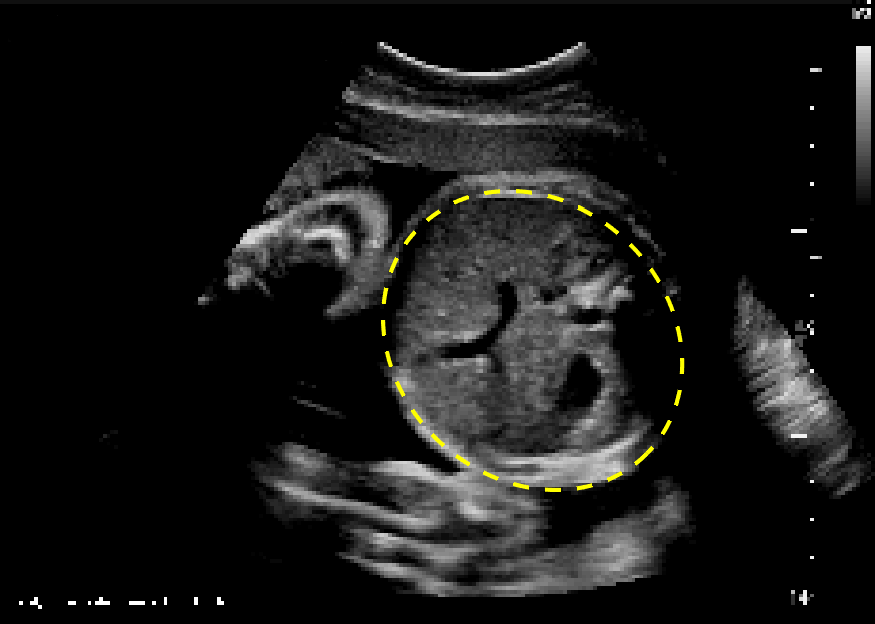}}
\subfigure[]
{
\includegraphics[width = .23\textwidth]{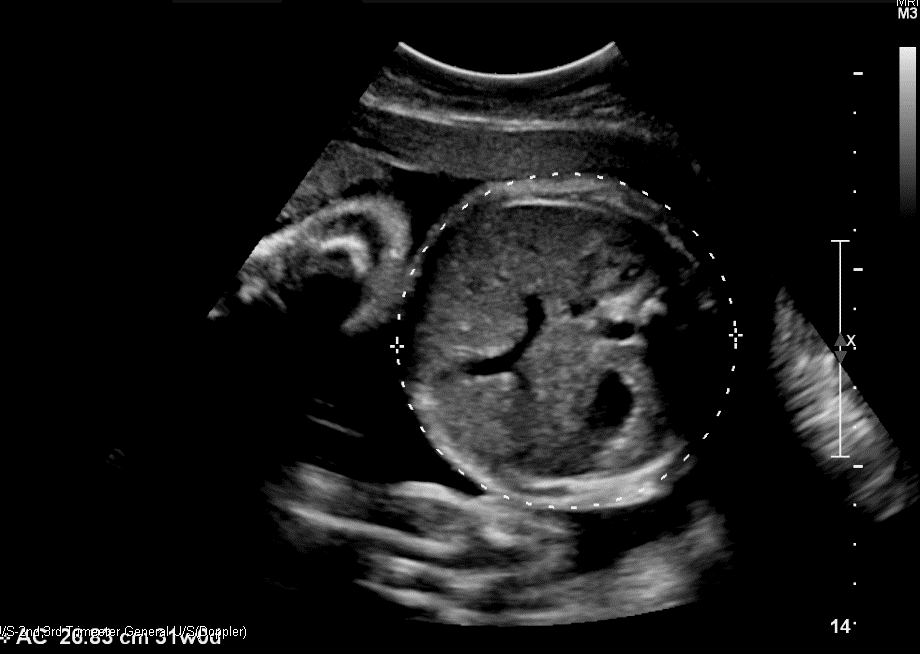}}
\caption{ The result for AC measurement. By applying the proposed CNN to (a) and (e), corresponding semantic segmentation images, (b) and (f), are obtained, respectively. (c) and (g) are the AC localization by the proposed Hough transform-based method. The placement of calipers are similar with experts' caliper placement in (d) and (h). \label{fig:results}}
\end{figure*}
For the assessment of AC measurement, the ultrasound images labelled as true axial planes by the experts with AC measurement were used to perform semantic segmentation using the proposed CNN.
At every anechoic point in the images, image patches corresponding to the normal- and wide-views, and ultrasound propagation direction were used as the input of the proposed CNN.
As described in Fig. \ref{fig:results}, the classification results for anechoic points in a given ultrasound image are represented as color maps, whose red, green, blue, and gray colors correspond to SB, UV, AF, and SA, respectively.

{ 
Since it could be very inefficient to choose candidate ellipses among all possible ellipses, we should filter out worthless ellipses.
For example, ellipses which overlap amniotic fluid region or have an abnormal ratio between the major and minor axes have a low possibility to be selected as candidates.
In order to make clear criteria, we used 26 true abdominal images with AC annotations and length from the training images.
From the true abdominal images, $0.6$ is heuristically chosen as the lower bound of the ration between the major and minor axes of ellipse which fits to AC annotation.
With the criteria, 25 candidates of ellipsoidal contours were selected from a single AF region image extracted from a semantic segmentation image and the median of parameters were chosen.
Although AC contours were localized abdominal regions well by using the lower bound of the ratio, the detected contours tend to be larger than contours annotated by the experts as described in Fig. \ref{fig:results}.
By comparing the two AC measurement results with experts' measurement with the training images, we decided to multiply $0.9$ to adjust the automatic measurement.

The ellipse detection were tested with the provided 40 true abdominal images with AC annotation from the test images.
Although false SB and UV regions are observed in AF regions and fetal abdomen is not fully surrounded by AF, the proposed Hough transform-based approach segments fetal abdominal regions.
We compared the AC estimation from accepted ultrasound images between the experts and our method.
Some comparison of abdominal contours selected by the experts and ellipsoidal contour are plotted in Fig. \ref{fig:results}.

We evaluated the performance of AC measurement with dice similarity metric
\begin{equation}
  d(O_\text{GT}, O_\text{AC}):= \f{2|O_\text{GT}\cap O_\text{AC}|}{|O_\text{GT}|+|O_\text{AC}|}
\end{equation}
where $O_\text{AC}$ is the abdomen region obtained from our AC estimation and $O_\text{GT}$ is the ground truth abdomen region delineated by the doctors.
This similarity metric $d(\cdot,\cdot)$ explains how the ground truth and detected region are close to and overlapped with each other.
The dice similarity was ${85.28\pm10.08}$\% for 56 cases whose AC measurement was given by just one expert.
For the top 80\% dice similarity, the score was ${89.19\pm4.2}$\%.
}

\subsection{Acceptance Check}
{  To train a CNN for plane acceptance check, we used 265 true and false transabdominal plane images from images of the training cases.}
Training and test sets for the acceptance check CNN consist of 209 and 56 cases of the annotated images, respectively.
For each case in the training and test sets, semantic segmentation was performed and fetal abdominal region was localized by using the proposed Hough transform-based approach.
Based on the localization, the semantic image was cropped and rescaled as mentioned above.
To augment our training set, the rescaled image was rotated with every 20 degrees from 0 to 340 degree and mirrored.
For training, ADAM \cite{Kingma2014} and dropout \cite{Srivastava2014} were applied, too.
After the training, a threshold level to accept a true axial plane is determined to maximize test accuracy of the proposed acceptance check CNN by using the test set.

For the performance evaluation, 105 transabdominal images among the annotated ultrasound images {  from images of the test cases} were used for evaluation. 
For the valuation, we compared acceptance check results among the 2 experts and the CNN by using the accuracy $A$:
\begin{equation}
 A = \frac{N_{tp}+N_{tn}}{N_{tp}+N_{tn}+N_{fp}+N_{fn}}
\end{equation}
where $N_{tp}$, $N_{tn}$, $N_{fp}$, and $N_{fn}$ are the number of true positive, true negative, false positive, and false negative, respectively.
As shown in Table \ref{tbl:AcceptanceCheck}, the accuracy of our acceptance check results are 0.809 and 0.771 with the expert 1 and expert 2, respectively while the accuracy between the two experts is 0.905.

\begin{table*}
\caption{Confusiion matrix for the acceptance check among the experts and the CNN for acceptance check \label{tbl:AcceptanceCheck}}
\centering

  \begin{tabular}{c c | c c | c}
  &         & \multicolumn{2}{c|}{Expert \# 1}& \multirow{2}{*}{Total} \\
  &        &False & True   \\\hline
   \multirow{2}{*}{Expert \# 2}&False     &    75   & 2 & 77\\
                         &True   &  8     & 20 & 28 \\\hline
  \multicolumn{2}{c|}{Total} & 83 & 22 & 105
 \end{tabular}
  \begin{tabular}{c c | c c | c}
  &         & \multicolumn{2}{c|}{Expert \# 1}& \multirow{2}{*}{Total} \\
  &         &False & True   \\\hline
   \multirow{2}{*}{CNN}&False     &    69   & 6 & 75\\
                         &True    &  14     & 16 & 30 \\\hline
  \multicolumn{2}{c|}{Total} & 83 & 22 & 105
 \end{tabular}
  \begin{tabular}{c c | c c | c}
  &         & \multicolumn{2}{c|}{Expert \# 2}& \multirow{2}{*}{Total} \\
  &           & False & True\\\hline
   \multirow{2}{*}{CNN}&False    &    64   & 11 & 75\\
                         &True    &  13     & 17 & 30 \\\hline
  \multicolumn{2}{c|}{Total} & 77 & 28 & 105
 \end{tabular}

\end{table*}

\section{Discussion and Conclusion}

Although CNN showed good performance in image recognition recently, it requires the collection of a large amount of training data to achieve satisfactory pixel-wise classification results. Unfortunately, owing to the limitation of gathering clinical data, it is difficult to collect sufficient data to guarantee satisfactory classification for various cases of ultrasound images. If a CNN is trained only with image data owing to the lack of physical characteristics, the number of training data required increases. We attempted to evade this problem by designing modality-specific structured CNN and obtained notable improvement in the training performance.

We also used CNN technique to evaluate the suitability of a selected image for proper AC measurement, where the suitability check was performed by analyzing anatomical configuration of the SB and UV regions in semantic segmentation images.
With 3D ultrasound imaging systems, this CNN method can be used to select the best abdominal biometry plane from volume data.
Other approaches for the suitability evaluation are reported in the literatures \cite{Hao2015,Ni2014,Kumar2015}.

\begin{figure*}[h!]
\centering
\subfigure[]
{
\includegraphics[width=.23\textwidth]{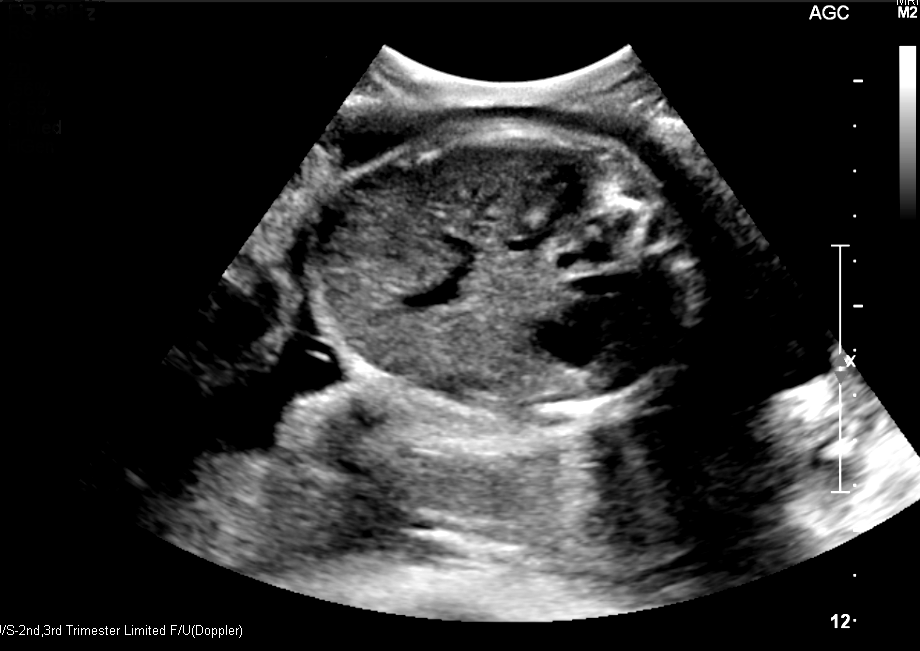}}
\subfigure[]
{
\includegraphics[width=.23\textwidth]{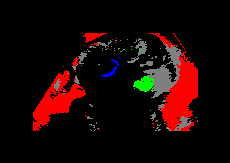}}
\subfigure[]
{
\includegraphics[width=.23\textwidth]{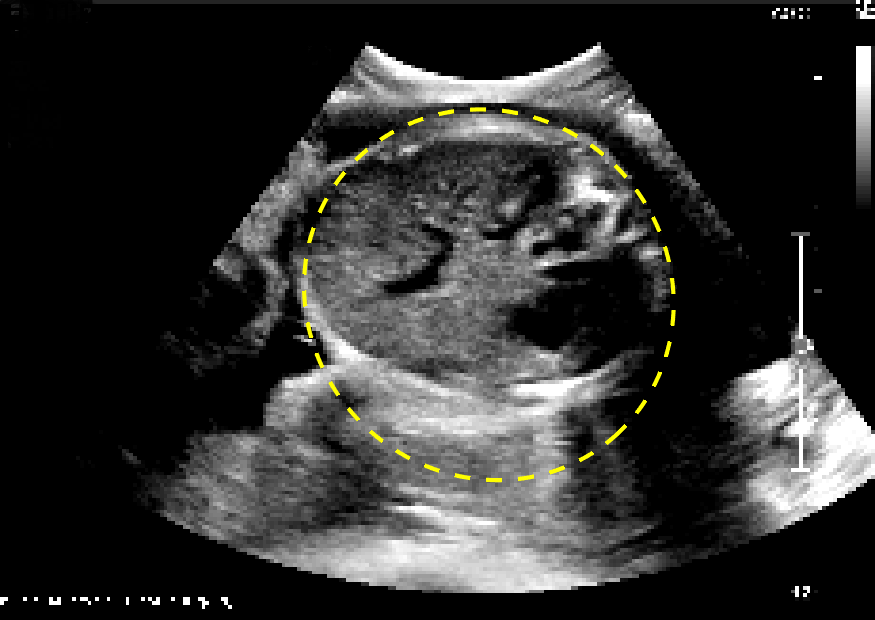}}

\subfigure[]
{
\includegraphics[width=.23\textwidth]{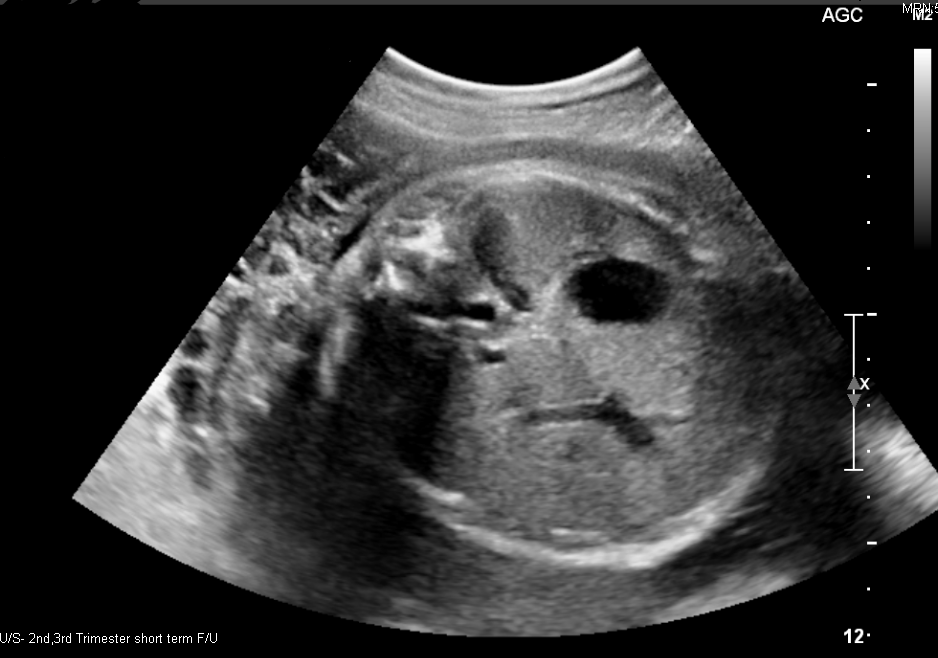}}
\subfigure[]
{
\includegraphics[width=.23\textwidth]{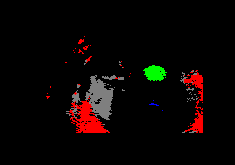}}
\subfigure[]
{
\includegraphics[width=.23\textwidth]{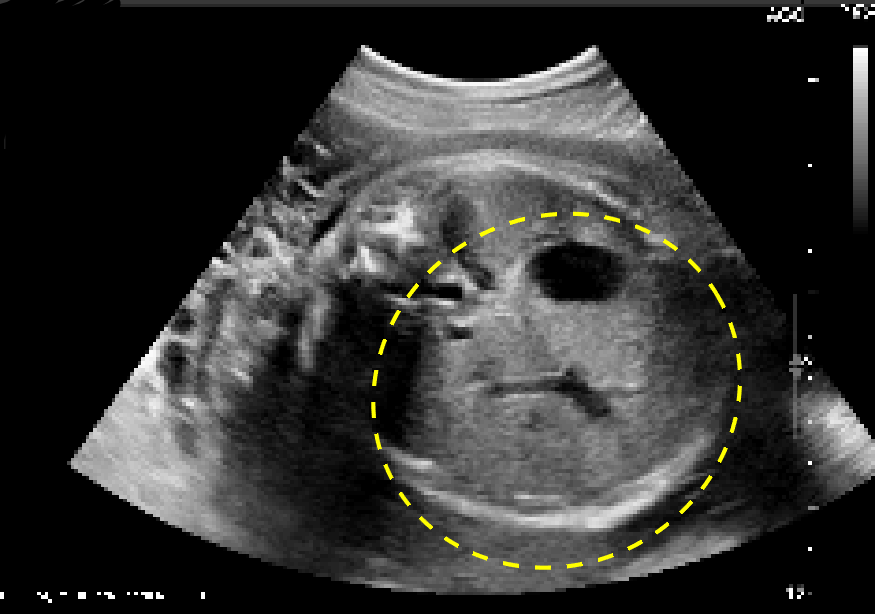}}
\caption{Cases whose AC fittings are inapproparate. In the first row, the fitted ellipse in (c) is properly localized but has bigger shpae due to the lack of abdominal boundary information along the minor axis direction.  In the case of the second row,  a lack of AF region on the left part causes underestimate of AC even though SA region in fetal abdomen is classified. \label{fig:unsuitableAC}}
\end{figure*}

The proposed method has a room for further improvements.
First, abdominal region detection should be refined because it does not influence only AC estimation but also localization of the abdominal region which is used for the acceptance check.
Fig. \ref{fig:unsuitableAC} indicates cases that detected ellipses by our method are insufficient to be accepted although SA in fetal abdomen is classified well.
In both cases in Fig. \ref{fig:unsuitableAC}, detected ellipses have relatively acceptable position but a lack of AF region causes such inaccurate caliper localization.
As explained, in clinical situations, it could not be guaranteed that sufficient anechoic points appear in AF region of ultrasound image for accurate and stable ellipse detection.
Therefore, it would be desirable to combine our method with a supplementary method {  or develop an advanced CNN to detect the ribs and spine where they are known to be crucial for acceptance check and AC fitting}.

Second, although we augmented the given  true and false abdominal plane images, the numbers of the true and false images were not balanced to guarantee a balanced performance for the true and false cases.
For false abdominal images, our acceptance check performed well while the accuracy for true abdominal planes is lower than the one for the false planes.
And the given true images are not sufficient to represent features of true abdominal planes.

{ 
Additionally, established architectures can be adopted to a part of proposed CNN architecture, such as U-net.
Due to a limited memory of our computing environment, we had a difficulty in sharing and training the established architectures.
With a sufficient computing environment, the performance could be improved by sharing the established architectures and their pre-trained filters, which is called "domain-transferred" deep CNN \cite{Hao2015}.
}

In our experiments, because of technical difficulties, only fetal ultrasound images were provided without probe geometry which are available when the method is implemented into a ultrasound system.
Due to the absence of the information, the performance of our framework could decrease.
In our experiments, in order to evaluate ultrasound propagation direction, we assumed that the probe is located at a certain point over the image (Fig. \ref{fig:dirFilter}) and the position was applied to all provided images even though the images have different imaging range.

{ 
In our results, there is no performance comparison to existing methods of automatic AC measurement because none of existing methods provides a stable performance and quantitative results under the similar experimental environmental with us.
We may refer to \cite{Rueda2014} for quantitative results for other parts, such as head and femur.
}

In summary, we proposed a method for automatic estimation of AC from ultrasound images.
This method shows good performance in most cases with relatively small number of training data.
This suggests that machine learning might find a breakthrough in the medical imaging field by focusing on developing modality-specific structures of CNN.
Even though the proposed method shows some limitations in cases of oversized fetuses and images highly corrupted by shadowing artifact, {  we expect that the proposed method of automated AC measurement contribute to measure accurate AC leading to estimate fetal weight accurately as well as to decrease operator dependency of AC measurement. Furthermore, our method will be helpful for artificial intelligence technique of automated measurement on ultrasonography in addition to current automation techniques.}

\section{Acknowledgements}
This work was supported by the National Institute for Mathematical Sciences (NIMS) grant funded by the Korean government (No. A21300000) and the National Research Foundation of Korea (NRF) grant 2015R1A5A1009350 and 2017R1A2B20005661.
%

\end{document}